\documentclass[conference]{IEEEtran}
\IEEEoverridecommandlockouts
\usepackage{cite}
\usepackage{multirow}
\usepackage{booktabs}
\usepackage[normalem]{ulem}
\usepackage{amsmath,amssymb,amsfonts}
\usepackage{listings}
\usepackage{algorithm}
\usepackage[noend]{algpseudocode}
\usepackage{caption}
\usepackage{subcaption}
\usepackage{float}
\usepackage{graphicx}
\usepackage{textcomp}
\usepackage{xcolor}
\usepackage{calc}
\usepackage{hyperref}
\usepackage{tikz}
\usepackage{textcomp}
\newlength\MAX  
\newlength\OFF  

\usepackage{algpseudocode}
\def\BibTeX{{\rm B\kern-.05em{\sc i\kern-.025em b}\kern-.08em
    T\kern-.1667em\lower.7ex\hbox{E}\kern-.125emX}}
\begin{document}
\newcommand{\pluseq}{\mathrel{{+}{=}}}
\newcommand{\minuseq}{\mathrel{{-}{=}}}

\title{MoRE-LLM: Mixture of Rule Experts Guided by a Large Language Model
}

\author{\IEEEauthorblockN{Alexander Koebler\IEEEauthorrefmark{1}\textsuperscript{,}\IEEEauthorrefmark{2}, Ingo Thon\IEEEauthorrefmark{2}, Florian Buettner\IEEEauthorrefmark{1}\textsuperscript{,}\IEEEauthorrefmark{3}}
\IEEEauthorblockA{\IEEEauthorrefmark{1} Goethe University Frankfurt, Frankfurt, Germany}
\IEEEauthorblockA{\IEEEauthorrefmark{2} Siemens AG,
Munich, Germany}
\IEEEauthorblockA{\IEEEauthorrefmark{3} German Cancer Research Center (DKFZ), Heidelberg, Germany}
\IEEEauthorblockA{Email: alexander.koebler@gmx.de, ingo.thon@siemens.com, florian.buettner@dkfz-heidelberg.de}
}

\maketitle

\begin{abstract}
To ensure the trustworthiness and interpretability of AI systems, it is essential to align machine learning models with human domain knowledge. This can be a challenging and time-consuming endeavor that requires close communication between data scientists and domain experts. Recent leaps in the capabilities of Large Language Models (LLMs) can help alleviate this burden. In this paper, we propose a Mixture of Rule Experts guided by a Large Language Model (MoRE-LLM) which combines a data-driven black-box model with knowledge extracted from an LLM to enable domain knowledge-aligned and transparent predictions. While the introduced Mixture of Rule Experts (MoRE) steers the discovery of local rule-based surrogates during training and their utilization for the classification task, the LLM is responsible for enhancing the domain knowledge alignment of the rules by correcting and contextualizing them. Importantly, our method does not rely on access to the LLM during test time and ensures interpretability while not being prone to LLM-based confabulations. We evaluate our method on several tabular data sets and compare its performance with interpretable and non-interpretable baselines. Besides performance, we evaluate our grey-box method with respect to the utilization of interpretable rules. In addition to our quantitative evaluation, we shed light on how the LLM can provide additional context to strengthen the comprehensibility and trustworthiness of the model's reasoning process.
\end{abstract}
\section{Introduction}
\begin{figure}[tb]
\centering
\includegraphics[width=\columnwidth]{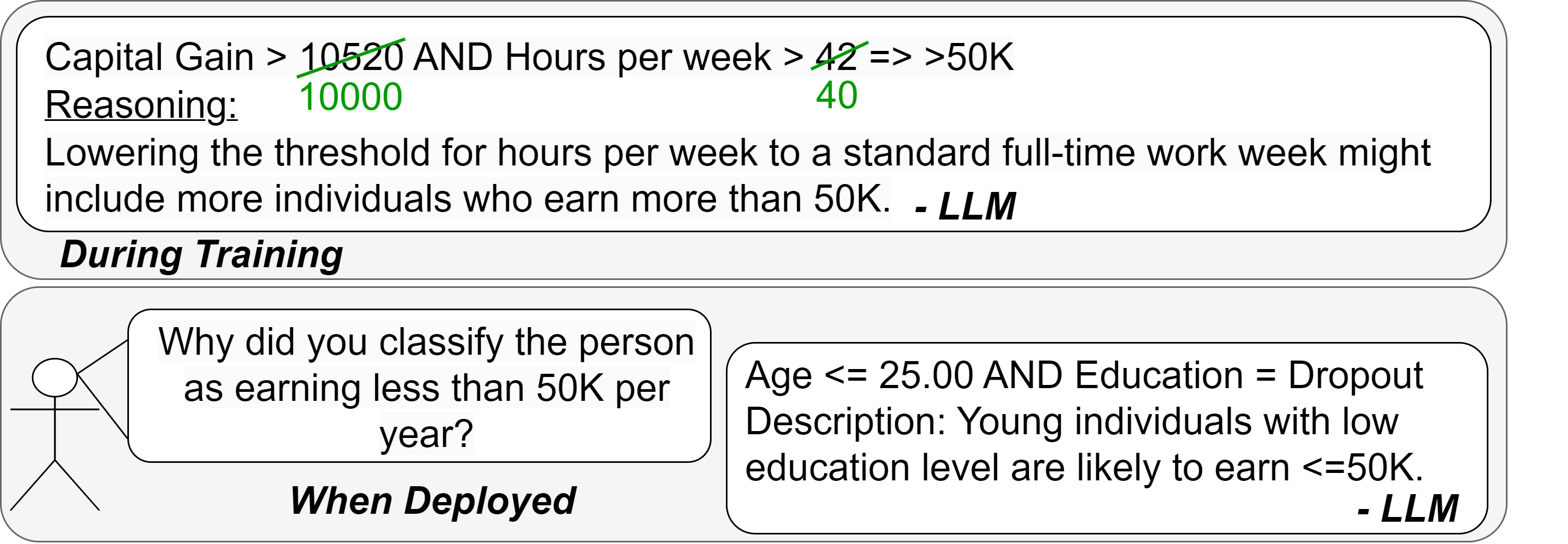}
\caption{In MoRE-LLM, the LLM is utilized in two steps of the model’s life-cycle. During training, it aligns discovered rules with domain knowledge, while during testing, insights generated by the LLM augment the model’s interpretability.} \label{fig:fig1}
\vspace{-0.5cm}
\end{figure}

Recent advances in the capabilities of Large Language Models (LLMs) \cite{achiam_gpt-4_2023} have opened up a multitude of new application areas. This is especially true for virtual assistance, where the human is kept directly in the loop. However, the uptake of these models in safety-critical or fully automated application areas is substantially slower. We see two main reasons for this. First, hallucinations in LLMs \cite{zhang_sirens_2023} can lead to non-factual outputs.
Second, general-purpose LLMs tend to require large computational resources to run.
On the other hand, the use of small, purely data-driven machine learning models in these applications is also subject to several challenges. Even if a human is not directly involved in every single decision process, the systems should allow for a human-on-the-loop setting that explains predictions as needed. Fulfilling this requirement is challenging due to the black-box nature of the deep learning models used, lacking interpretability. To address this issue, various post-hoc explanation methods have been proposed to describe the reasoning process of a black-box model \cite{guidotti_stable_2022, ribeiro_anchors_2018, ribeiro2016should}. However, these explanations only approximate the model's decision process. Moreover, they often reveal a misalignment between the decision process of the machine learning model and a human expert, which reduces the user's trust in the AI system. Recent works emphasize the importance of grounding both the machine learning models and the generated explanations in human domain knowledge \cite{decker2023thousand}.
With Mixture of Rule Experts Guided by a Large Language Model (MoRE-LLM) we propose the first framework that utilizes an LLM to guide a small task-specific model. MoRE is a Mixture of Experts (MoE) that combines a black-box model with a rule-based classifier to offer high-fidelity rule-based explanations for a subset of the input space. These rules serve as an interface for the LLM to align the reasoning process with domain knowledge during an iterative learning process. A data-driven gating model determines whether a rule should be used for a particular instance, taking into account potential hallucinations induced by the LLM that contradict empirical observations in the real-world training data. During the training phase, the LLM aligns the task specific model with domain knowledge by  refining and pruning rules;  during deployment, rules serve as explanations grounded in this domain knowledge and the LLM further enhances interpretability by providing additional context to the rules. The context being generated during training time alongside the rules removes the need for access to the LLM after the model is deployed. Figure \ref{fig:fig1} illustrates both ways in which the LLM facilitates the AI system in different steps of the ML life-cycle. The interaction between the MoRE model and the LLM is fully automated.
Taken together, we propose a novel approach that allows small task-specific models to benefit from the in-depth knowledge modern LLMs have acquired from extensive and diverse training data, whilst safeguarding against factual inaccuracies or ''hallucinations''.
The main contributions of our work can be summarized as:
\begin{itemize}
    \item We introduce a Mixture of Experts (MoE) based architecture to combine a black-box neural network model with a learnt rule set in a grey-box classifier, which is trained via end-to-end-optimization.
    \item We propose a novel approach to sample local rule surrogates of a black-box model using Anchors \cite{ribeiro_anchors_2018} and aggregate them in a white-box rule-based classifier.
    \item We re-anchor logical rules in domain knowledge via LLMs, while simultaneously safeguarding against factual inaccuracies through a learned gating function.
    \item We maximize the utilization of the rules without sacrificing predictive power in a highly non-convex constrained optimization setting by building on the Dynamic Barrier Gradient Descent (DBGD) \cite{gong_automatic_2021}.
\end{itemize}
MoRE-LLM utilizes the synergy between multiple yet rather separated research fields to facilitate the development of domain knowledge aligned and interpretable task-specific models. It can be considered a framework which also allows for future substitution of components such as the explanations method.
\section{Related Work}
When aiming for interpretable predictions, most approaches distinguish between two paradigms:  the use of post-hoc explanations on black-box models \cite{guidotti_stable_2022,ribeiro_anchors_2018,ribeiro2016should} or inherently interpretable models \cite{yang_learning_2021, yang_truly_2022}.
In both cases, it is essential that the generated explanations are comprehensible for the human user. Therefore, rule-based explanation methods \cite{guidotti_stable_2022, ribeiro_anchors_2018, sharma_maire_2020} have proven to be beneficial for tabular data sets. The Anchors method introduced in \cite{ribeiro_anchors_2018} extends the well established LIME \cite{ribeiro2016should} approach by generating rule based surrogate models that fit the prediction of a black box model in the proximity of the input sample. Additionally to explaining the prediction itself, the authors in \cite{guidotti_stable_2022} generate counterfactual rules indicating how the input must change to lead to a different outcome. Both methods approximate the decision process underlying a given prediction and do not guaranteeing full fidelity. The method introduced in \cite{setzu_glocalx_2021} is close in spirit to our rule set learning approach as the authors aggregate local rule explanations to a global surrogate model eventually substituting the original black-box model completely. However, the method does not consider a hybrid combination of both models.
Independent of local rule explanation, multiple works try to combine interpretable and black-box approaches \cite{wang_gaining_2019, pradier_preferential_2021, gong_automatic_2021}.
In \cite{wang_gaining_2019} the authors propose a method to build a decision rule set to substitute the prediction of a black box model for a subset of the input data. However, the method does not consider a gating model which would allow to refuse assigning a sample to the rule set even if it would yield a prediction, therefore, it can not account for low quality predictions of the rules. However, this is essential to handle LLM generated rules subject to potential hallucinations.
Preferential Mixture-of-Experts \cite{pradier_preferential_2021} aims to allow for providing human rules alongside a black box model using a Mixture of Experts (MoE) approach. In their work the authors introduce a constrained optimization objective to prefer the interpretable model as long as a predefined performance constraint is meet. We adopted this constraint and extended on their approach by substituting the suggested optimization methods, with the Dynamic Barrier Gradient Descent (DBGD) introduced in \cite{gong_automatic_2021} to allow for non-convex constrained optimization problems due to the use of deep learning models in the MoE. None of the mentioned works does consider the utilization of LLMs for knowledge-alignment of extracted concepts or rules. 
\section{Problem Setting}
For the introduced approach, we suppose a supervised classification setting with labeled training data $\mathcal{D} = \{(x_n, y_n)\}^N_{n=1}$ consisting of $N$ input samples $x_n \in \mathbb{R}^d$ and corresponding targets $y_n$. Further we assume access to a test dataset $\mathcal{D}_t = \{(x_n, y_n)\}^M_{n=1}$ for evaluation.
When training a classifier $f_{\theta^*}: \mathcal X \rightarrow \mathcal Y$ with parameters $\theta^*$ we can measure the performance of this black-box model on the training set by an appropriate loss function $\mathcal{L}_{task}(f_{\theta^*}) = \sum^N_{n=1} l_{task}(f_{\theta^*}(x_n),y_n)$. Here, we use a cross-entropy loss function. In the considered setting, $f$ might be a black-box model, e.g. a Multi-Layer Perceptron (MLP), leading to the predictions $f_{\theta^*}(x_n)$ not being interpretable.
\section{Methodology}
\begin{figure}[tb]
\centering
\includegraphics[width=\columnwidth]{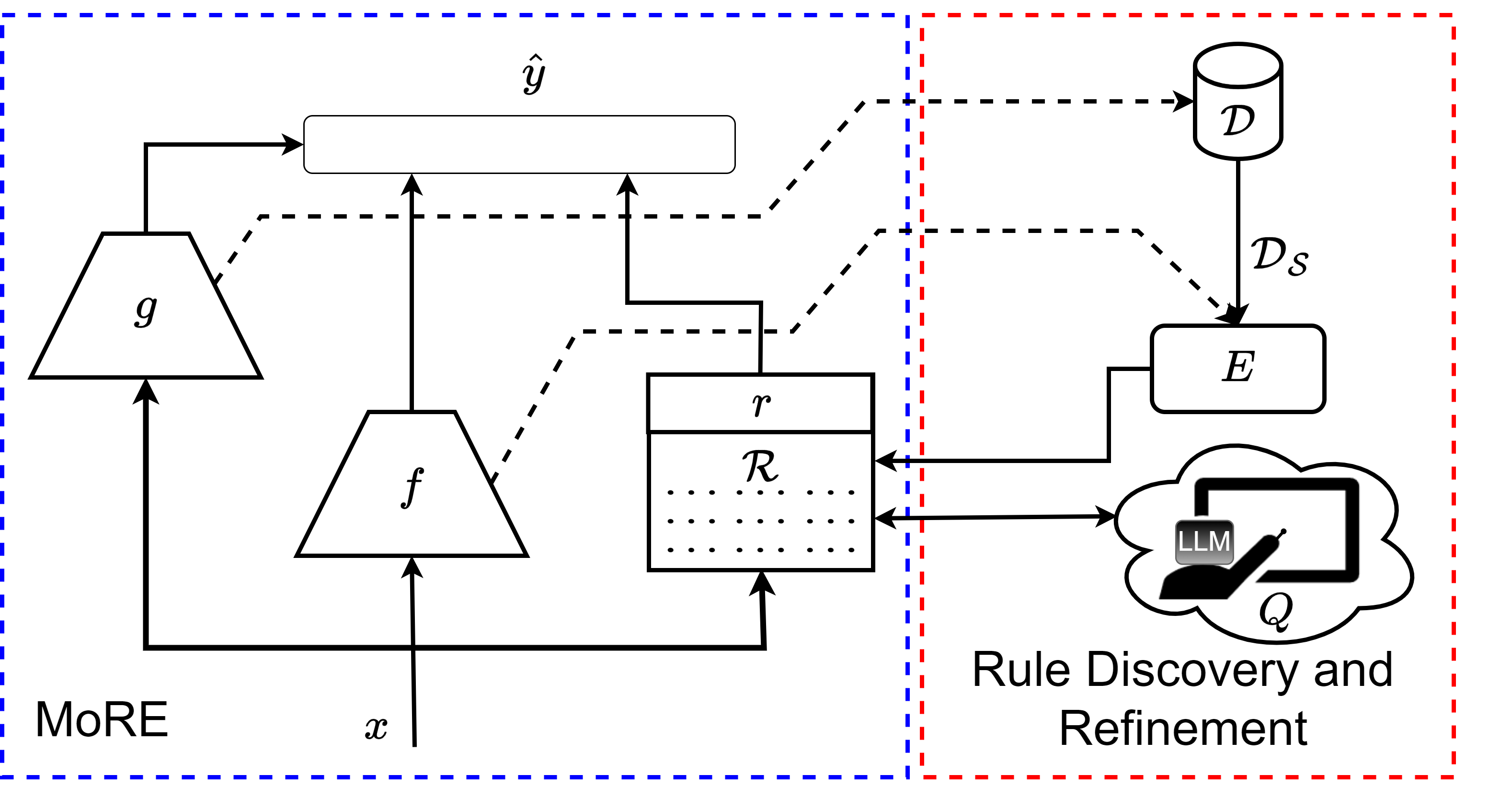}
\caption{Overall MoRE-LLM architecture. The elements encapsulated by the blue box consisting of gating model $g$, black-box classifier $f$ and the rule-based classifier $r$ including rule set $\mathcal{R}$ are required during test time. The training set $\mathcal{D}$, the large language model $Q$ and the explainer module $E$ in the red box are only necessary during training time.} \label{fig:method}
\vspace{-0.5cm}
\end{figure}

In the following section, we elaborate on our proposed framework, which consists of the MoRE architecture illustrated in Figure \ref{fig:method}, and an iterative training procedure summarized as follows:
In an initialization step, we train the black box model $f$ in an unconstrained manner. Afterward, we enter a loop by generating rules $R$ as local surrogates of the current model $f$. Next, an LLM $Q$ is queried to adapt the discovered rules, along with potential rules from previous iterations, based on domain knowledge. These adapted rules are then used for classification in a rule-based classifier $r$. In a constrained optimization step, the black-box model $f$ and a gating model $g$ are optimized such that the rule model $r$ substitutes $f$ for predictions as much as possible while maintaining the performance of the black-box model $f$ trained in the initial step. The next iteration continues by discovering new rules for regions in the input space that have not yet been assigned by the gate $g$ to the rule model $r$.
\paragraph{Mixture of Rule Experts}
With the Mixture of Rule Experts (MoRE), depicted in Figure \ref{fig:method}, we introduce a rule predictor $r$, which relies on a rule set $\mathcal{R}$, as well as a gating model $g_{\omega} = (g^1, g^2)$, with parameters $\omega$ and two outputs $g^1$ and $g^2$, alongside the black-box model $f_{\theta}$.
The rule set $\mathcal{R} = \{R^1, R^2, \dots , R^U\}$ consists of $U$ rules which in turn are a set of predicates. The rule-based predictor $r_\mathcal{R}(x) = (c^1, c^2, \dots, c^C)$ outputs a one-hot vector having the same shape as $f_{\theta}(x)$ which is $1$ for the predicted class and $0$ for all other $C - 1$ classes. During the iterative discovery process of rules $R^u$ as local surrogates, elaborated in detail later, a corresponding training sample used for generating the local surrogate $x^u$ is assigned and stored alongside every rule. However, rules $R^u$ generalize beyond the single samples $x^u$, thus for predicting $r_\mathcal{R}(x)$ the rule $R^u$ is used where $x^u$ is closest to $x$ and $R^u$ does classify $x$. To express an abstain if no $R^u$ classifying the provided instance is given, all elements $c^i$ of $r_\mathcal{R}(x)$ are set to $0$. Taken together where $g_{\omega}$ weighs the predictions of $r_\mathcal{R}$  and $f_{\theta}$, MoRE yields the output $\hat y$ for input $x$ as:
\begin{align*}
\hat y = 
\begin{cases}
    g^1_{\omega}(x) \cdot f(x) + g^2_{\omega}(x) \cdot r_\mathcal{R}(x),& \text{if } \sum\limits_{c^i \in r_\mathcal{R}(x)}c^i = 1\\
    f(x),              & \text{otherwise}
\end{cases}
\end{align*}
To optionally guarantee a discrete assignment of input instances either to the interpretable rule-based model $r_\mathcal{R}$ or the black-box classifier $f_{\theta}$ during inference, $g_{\omega}(x)$ can be one-hot encoded before being applied. 

\paragraph{Constrained Gate Optimization}
To encourage the utilization of the interpretable rule-based predictor $r_\mathcal{R}$, we introduce an auxiliary interpretability loss $l_{int}(x) = -log(g^2_{\omega}(x))$,
which given the softmax in the gating model $g_{\omega}$ maximizes the assignment of instances $x$ to $r_\mathcal{R}$ and minimizes the assignment to $f_{\theta}$.
We aim that the gate $g_{\omega}$ assigns as many predictions as possible to the interpretable rule set whilst maintaining similar predictive performance $\mathcal{L}_{task}(f_{\theta^*})$ as the black-box model $f_{\theta^*}$ trained in an unconstrained manner in the initialization step. More precisely, the loss of the resulting grey-box model should hold $\mathcal L_{task}(g_{\omega},f_{\theta},r_\mathcal{R}) \leq (1+\epsilon) \mathcal{L}_{task}(f_{\theta^*})$. Thus, our initial constrained optimization objective can be formalized as 
\begin{align*}
    \min_{\omega,\theta} &\sum^N_{n=1} l_{int}(x; \omega) \quad \\ &s.t. \quad \mathcal L_{task}(g_{\omega},f_{\theta},r_\mathcal{R}) \leq (1+\epsilon) \mathcal{L}_{task}(f_{\theta^*}) \, .
\end{align*}
Considering, that we want to allow the classification model $f_{\theta}$ to specialize on areas which are not covered by the rules or which are covered by rules that cannot provide sufficient accuracy, we optimize for both objectives simultaneously.
The simplest approach for optimizing this two goals and handling the constraint is given by a linear combination of both objective functions
\begin{align*}
    \min_{\omega, \theta} \sum^N_{n=1} (l_{int}(x_n; \omega) + \lambda l_{task}(x_n; \omega, \theta)) \, .
\end{align*}
For the simple linear combination of both objectives, the fulfilment of the constraint is highly dependent on the weight coefficient $\lambda$ and neither objective can be prioritised.
To alleviate this issue, \cite{pradier_preferential_2021} utilize a log-barrier gradient descent and a projected gradient descent approach while using logistic regression models for the classification and gating model.
Since we aim to develop a non-linear model with high predictive power, we want to allow for the usage of a neural networks for both $f_{\theta}$ and $g_{\omega}$. However, this implies that we have to solve a constrained highly non-convex optimization problem, which cannot be effectively solved via a relatively straight-forward log-barrier gradient descent. Instead, we extend the Dynamic Barrier Gradient Descent (DBGD) method introduced in \cite{gong_automatic_2021}. DBGD promises to allow for optimizing a secondary objective within the optimal set of a first objective. For this, the authors introduce a dynamic adaptive combination coefficient $\lambda_t$ that weighs the sum of the gradient resulting from both objectives in every optimization step $t$. We adapt this approach to our optimization problem and consider the optimization of the interpretability loss $l_{int}(x; \omega)$ as our secondary objective which should be optimized if the constraint on the classification performance $l_{task}(x; \omega, \theta)$ is fulfilled. Note, that the model $f_{\theta}$ does not or only marginally influence both loss functions in cases where the rule-based predictor $r$ is assigned, leading to vanishing gradients for $\theta$. Thus, we only apply the constrained optimization of $l_{int}(x; \omega)$ to the gating model in cases where rules are available and simultaneously optimize $f_{\theta}$ using $l_{task}(x; \omega, \theta)$ for all samples. Our proposed optimization procedure is described in detail in Algorithm \ref{const_op}.
\begin{algorithm}
  \caption{Optimization}\label{const_op}
  \begin{algorithmic}[1]
    \Procedure{optimization}{$f_{\theta}, g_{\omega}, \mathcal{D}, \mathcal{L}_{task}(f_{\theta^*}), \epsilon, \eta$}
      \For{\texttt{epoch e}}
        \For{\texttt{batch} $b \in \mathcal{D}$}
                \State Calculate $\nabla l_{task}(\theta)$ for $b$
                \State $\theta \gets \theta + \eta \cdot \nabla l_{task}(\theta)$ \Comment{Learning rate $\eta$}
                \State $\mathcal{I} = 0$, $\mathcal{T} = 0$
                \For{\texttt{instance} $i \in b$}
                    \If{$\sum r_{\mathcal{R}}(i)>0$} \Comment{Do rules apply?}
                        \State Calculate $\nabla l_{int}(\omega)$, $\nabla l_{task}(\omega)$ for $i$
                        \State $\mathcal{I} \pluseq \nabla l_{int}(\omega)$, $\mathcal{T} \pluseq \nabla l_{task}(\omega)$
                    \EndIf
                \EndFor
                \State $\phi = \min (\alpha (\mathcal{L}_{task}(f_{\theta}, g_{\omega}, r_{\mathcal{R}})-(1+\epsilon)\mathcal{L}_{task}(f_{\theta^*}), \beta ||\mathcal{T}||^2)$  \Comment{Here: $\alpha = \beta = 1$}
                \State $\lambda_t = \max (\frac{\phi - \mathcal{I}^T\mathcal{T}}{||\mathcal{T}||^2}, 0)$ \Comment{Adaptive coefficient}
                \State $\omega \gets \eta(\mathcal{I} + \lambda_t \mathcal{T})$\Comment{Constrained update of $g_{\omega}$}
            \EndFor         
      \EndFor
    \EndProcedure
  \end{algorithmic}
\end{algorithm}

\paragraph{Iterative Rule Discovery}
We introduce an iterative rule discovery approach utilizing the models $g_{\omega}$ and $f_{\theta}$ in our proposed MoRE architecture to guide the rule generation process. Thereby, we aim to emphasis two things. First, by generating rules $R^u$ only for areas which are previously assigned to $f_{\theta}$ by $g_{\omega}$ considering the performance constraint, we focus the rule discovery on areas where $r_{\mathcal{R}}$ is currently outperformed by $f_{\theta}$. Thus, we efficiently use our available budget for the number of rules and iterations to shrink the number of samples not assigned to $r_{\mathcal{R}}$.\newline
Second, rather then generating the rules to achieve an optimal global coverage and accuracy, we use an explainer module $E(x,f)$ generating local rule surrogates that follow the classifier $f_{\theta}$ as close as possible. We already know, that $f_{\theta}$ outperforms existing rules in that area to a degree violating our preset performance constraint. Since the new rules approximate the local decision boundary of the model $f_{\theta}$, this leads to a similar local performance and should substitute $f_{\theta}$ in a following optimization step.\newline
We use the Anchors approach introduced by Ribeiro et al. \cite{ribeiro_anchors_2018} to yield the local surrogate rules. This generates a single rule $R^u = E(x^u,f)$ for an input sample $x^u$. The rule approximates the performance of the black-box model $f_{\theta}$ up to a pre-set relative accuracy threshold $\tau$ within the proximity region of $x^u$.
In every iteration, a subset $\mathcal{D}_S \subset \mathcal{D}$ of the training data set is sampled for which rules are generated and appended to the rule set $\mathcal{R}$. To support new rules to increase the rule coverage without being redundant, the samples $x^u \in \mathcal{D}_S$ should fulfill two conditions.
First, all samples should be allocated to the model $f_{\theta}$ by the gate $g_{\omega}$. Second, out of this set a subset of length $B$ is chosen according to a mix of two sampling strategies. First, to \textit{exploit} areas where the classifier $f_{\theta}$ is highly certain about the prediction, we sample the examples in $\mathcal{D}_S$ with the lowest output entropy. Second, to support $f_{\theta}$ and \textit{explore} areas where it generates very uncertain predictions we sample examples with high output entropy. Whereas exploitation emphasises the trust in the correct prediction of $f_{\theta}$ in low entropy areas, exploitation pushes the responsibility for assessing the suitability of the rule to the gating model and to the adaptation of the rules via an LLM, as described in the following paragraph.\newline
Up to this point, the discovery process does not explicitly prevent the rediscovery of rules already contained in the rule set $\mathcal{R}$ or the generation of duplicates within a step. The creation of duplicate rules can be caused by sampling two very close $x^u$ resulting in the same local surrogate rule, or by the LLM simplifying rules and removing distinguishing predicates. To handle these cases, the duplicates consisting of the same predicates are removed after each rule discovery step.
\paragraph{LLM-based Rule Set Refinement}
For MoRE-LLM, we employ a rule refinement step to every iteration in order to align the rule set $\mathcal{R}$ with human domain and general knowledge encoded in the LLM $Q$. Thereby we regularize the rule discovery procedure with information outside of the training data, which might promise better generalization in real-world deployment scenarios.\newline
To automate the embedding of the LLM rule refinement step in the iterative rule discovery, we divide the task in a rule adaptation and a rule pruning step. This in combination with engineering appropriate prompts allows us to receive fixed form responses which can be parsed to automatically adapt the rule set $\mathcal{R}$ accordingly.
In the \textit{rule adaptation} step we allow the LLM to adapt all elements of the rules. However, the removal of an entire rule is not allowed in this step. The LLM can decide to remove predicates from rules. This is in line with some rule-pruning steps in conventional rule-learning algorithms and can counteract overfitting. Further, for numerical features, the LLM is allowed to adapt the operator or the threshold as well as for categorical features the output category. To prevent the model from coming up with own operators, categories or features which might not be included in the training data we have to strictly specify them in our prompt. There is no restriction for the model to keep the output class which would risk adaptations of the predicates yielding contradictions of the original output.

In the \textit{rule pruning} step the LLM is now allowed to specify rules which should be removed, even after they have been adapted. The reasons for removal can be rules still being over-specific, under-complex or contradicting domain knowledge. Additionally, given that the entire rule set is included in the prompt for every iteration, the LLM can also discover contradictions or high similarity between rules and initiate the removal for one of the rules. Besides stating which rules should be removed, the LLM is also asked to provide a reasoning for the decision. This reasoning offers an interface for a human domain expert to retrace the modeling process or even intervene if necessary.
\section{Experiments}
\paragraph{Experimental Setup}
In our conducted experiments both the gating model $g_{\omega}$ and the classifier $f_{\theta}$ share the same architecture. This is ether a MLP with two hidden layers of size $50$, or a Logistic Regression (LR) model. For the considered binary classification data sets, both $f_{\theta}$ and $g_{\omega}$ have two outputs followed by a softmax layer. For generating the rule surrogates we utilize the Anchors approach \cite{ribeiro_anchors_2018}. For rule generation we sample four samples according to the explore and four samples following the exploit sampling strategy. For the LLM $Q$ we use a GPT-4 \cite{achiam_gpt-4_2023} model. The slack parameter $\epsilon$ for the constraint is set to $0.1$ allowing for a $10\%$ loss increase in comparison to the unconstrained original model $f_{\theta^*}$.
We evaluate on three commonly used tabular data sets available via \cite{vanschoren_openml_2013}.
For quantitative evaluation, we compare our approach with six widely used methods for classification tasks on tabular data. These methods range from easily interpretable approaches such as RIPPER \cite{COHEN1995115} and CART \cite{Breiman1984ClassificationAR}, which provide direct access to the rule used for a particular prediction, over less interpretable tree ensembles such as AdaBoost \cite{FREUND1997119}, Gradient Boosted Decision Trees (GBDT) \cite{GBDT} and Random Forests (RF) \cite{ho_random_forests} up to a black-box MLP. Note that although tree ensembles offer some unique approaches in generating interpretations in the form of feature attributions \cite{lundberg2018consistent}, they are still often considered black-box approaches \cite{palczewska2014interpreting}. More details on the implementation as well as the used prompt templates are provided at: https://github.com/alexanderkoebler/MoRE-LLM
\paragraph{The LLM as a Teacher} \label{sec:teacher}
During \textit{rule adaptation} we can observe a number of different patterns. Among others those include adapting numerical values to align with domain knowledge or increase interpretability, see Figure \ref{fig:fig1}. Furthermore, if the rules contradict general or domain knowledge, the model swaps the output class or removes predicates which should not have an influence on the prediction.
\begin{figure}
\centering
\includegraphics[width=\columnwidth]{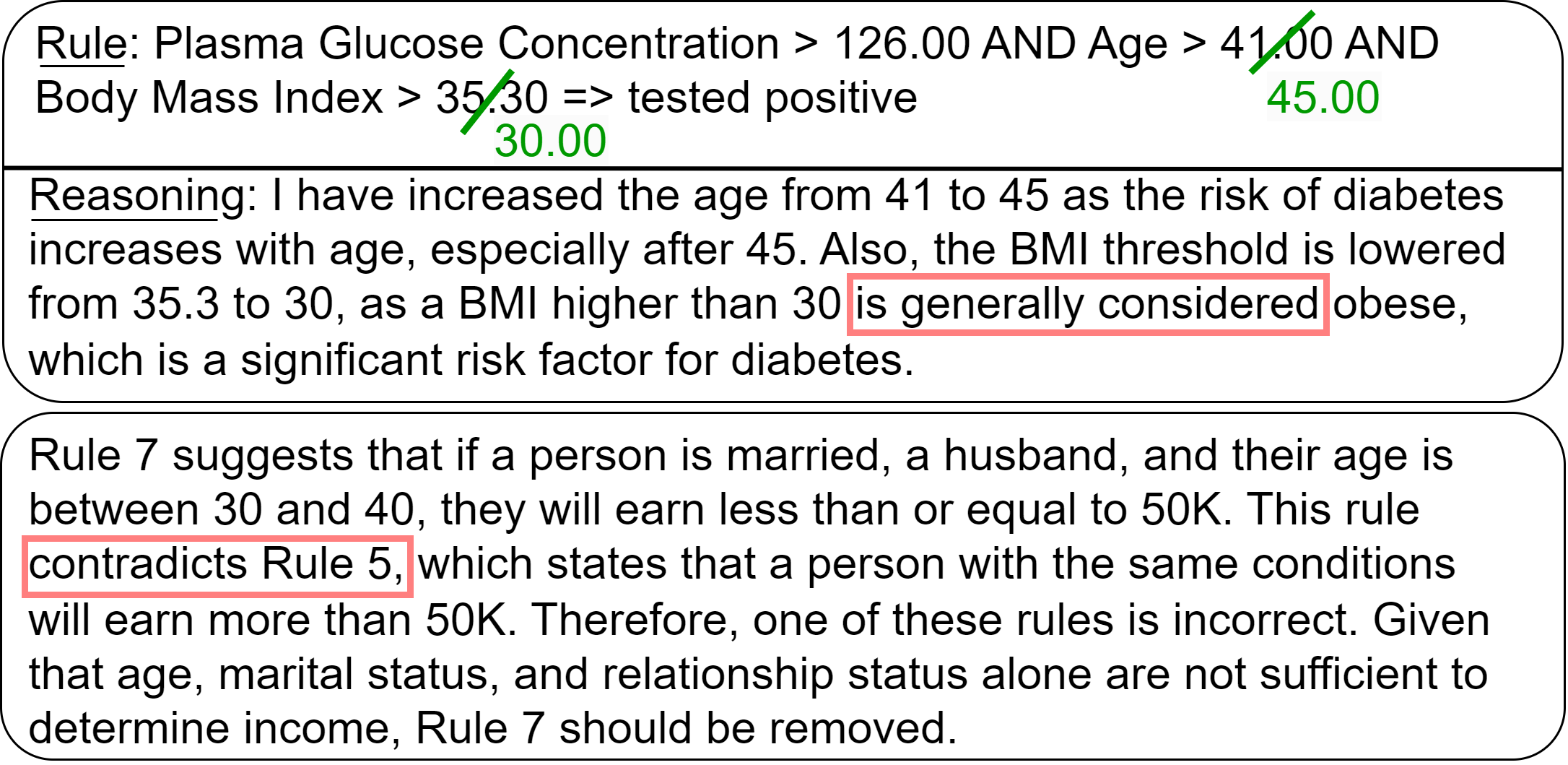}
\caption{Examples for LLM based rule refinement. The rule adaptation example on the diabetes dataset (top) relies on specific knowledge about health factors while the rule pruning example on the adult dataset (bottom) discovered contradictions in the context of the other rules.} \label{fig:adaptation}
\end{figure}
In the \textit{rule pruning} step on the other hand, we observed two main patterns. The LLM either removes a rule if it contradicts domain knowledge or one of the other rules. In the latter case the LLM argues to preserve the rule which is most aligned with domain knowledge, see Figure \ref{fig:adaptation}.

\begin{figure}[t!]
    \centering
    \begin{subfigure}[b]{0.5\columnwidth}
        \centering
        \includegraphics[width=\linewidth]{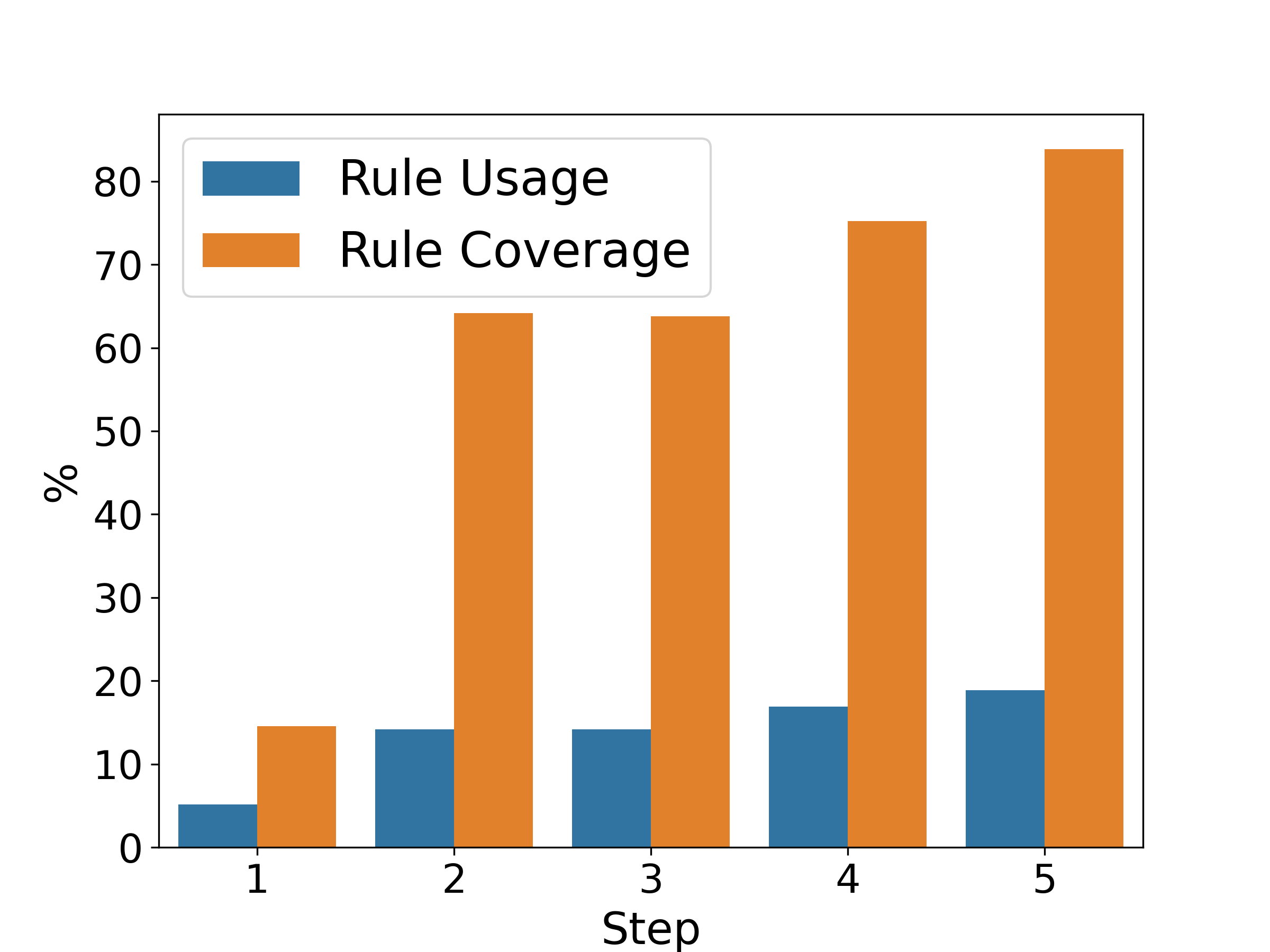}
        \caption{} \label{fig:steps_cov}

    \end{subfigure}%
    \begin{subfigure}[b]{0.5\columnwidth}
        \centering
        \includegraphics[width=\linewidth]{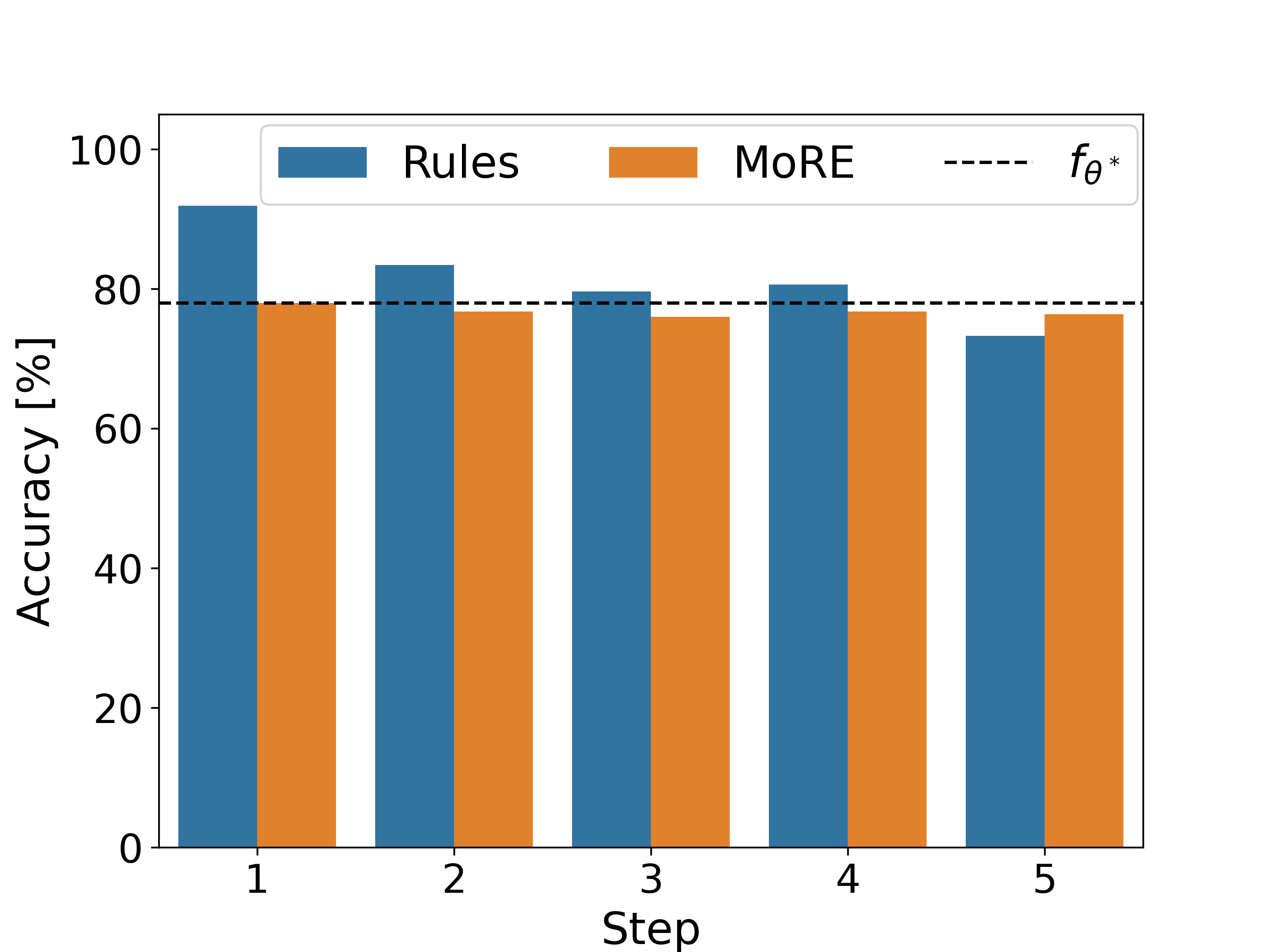}
        \caption{} \label{fig:steps_acc}
    \end{subfigure}
    \caption{Rule coverage and utilization (a) as well as test accuracy and accuracy of the generated rules (b) for MoRE-LLM (MLP) on a test set for the diabetes classification task across five consecutive steps.}
\end{figure}

An important side benefit of the LLM-generated justifications for adjusting or keeping a rule is that they can \textit{augment explanations} and enhance interpretability once the model is deployed. When a user requests an explanation for an instance associated with one of the rules, the LLM-generated description can be provided alongside the classification rule, as shown in Figure \ref{fig:fig1}. Since these descriptions are generated and stored with the rules during training, no access to the LLM is required during test time. For data instances where the black box model is used, regular lower fidelity post-hoc explanation methods like Anchors or LIME can be used to still provide some explanations. However, in these cases, it should be made transparent to the user that, unlike rule-covered instances, the model's prediction process might not exactly follow the provided explanations.
\paragraph{Performance and Rule Utilization}
To quantitatively evaluate the benefit of MoRE-LLM concerning interpretability, we utilize two metrics for the quality and utilization of the generated rule set. First, the rule $\text{Coverage} = \frac{1}{M}\sum_{x \in \mathcal{D}_t}\sum_{i=1}^C(r^i_{\mathcal{R}(x)})$ expresses how many of the $M$ data points in the test set $\mathcal{D}_t$ can be classified by the rule-based classifier $r_{\mathcal{R}}$, i.e., get assigned a class, whereas the $\text{Usage} = \frac{1}{M}\sum_{x \in \mathcal{D}_t}(g^2_{\omega}(x) > 0.5)$ indicates what number is actually assigned to $r_{\mathcal{R}}$ by the gating model $g_{\omega}$.
We consider an instance to be assigned to one of the models if the activation of the corresponding gate is above $0.5$.
In each iteration, we generate rules for instances sampled in regions not yet assigned to $r_{\mathcal{R}}$. This lack of assignment may occur because no rules cover a particular instance, or because the existing rules covering it fail to meet the performance constraint. Consequently, we anticipate that both rule coverage and usage will increase with each iteration. This trend is evident in Figure \ref{fig:steps_cov}.
We observe that rule coverage often significantly surpasses actual usage. This discrepancy suggests that the gating model deliberately avoids using rules if necessary to maintain adherence to the performance constraint. Our hypothesis gains further support from Figure \ref{fig:steps_acc}, which demonstrates that the model consistently maintains test performance close to that of the black-box model $f_{\theta^*}$ regardless of the rule coverage. Even when rule performance on covered examples decreases, the gating model effectively manages rule utilization to enforce the desired performance level.
\begin{table}[]
\centering
\caption{Comparison of the task loss as well as rule coverage (cov.) and usage (usg.) on test set after three iterations. We list the results for the MoRE approach with and without the LLM.}
\label{tab:res_util}
\resizebox{\columnwidth}{!}{
\begin{tabular}{@{}clllllllll@{}}
\toprule
\multirow{2}{*}{Method} & \multicolumn{3}{c}{adult}          & \multicolumn{3}{c}{g-credit}       & \multicolumn{3}{c}{diabetes}       \\ \cmidrule(l){2-10} 
                        & $\mathcal{L}_{task}$ & cov. & usg. & $\mathcal{L}_{task}$ & cov. & usg. & $\mathcal{L}_{task}$ & cov. & usg. \\ \midrule\midrule
LR             & 0.49 & -    & -    & 0.54 & -    & -    & 0.53 & -    & -    \\ \midrule
MoRE (LR)      & 0.50 & 0.77 & 0.02 & 0.55 & 0.72 & 0.06 & 0.56 & 0.89 & 0.20 \\ \midrule
MoRE-LLM (LR)  & 0.50 & 0.56 & 0.14 & 0.55 & 0.54 & 0.15 & 0.56 & 0.79 & 0.41 \\ \midrule\midrule
MLP            & 0.46 & -    & -    & 0.54 & -    & -    & 0.52 & -    & -    \\ \midrule
MoRE (MLP)     & 0.46 & 0.72 & 0.10 & 0.59 & 0.62 & 0.17 & 0.56 & 0.80 & 0.24 \\ \midrule
MoRE-LLM (MLP) & 0.47 & 0.66 & 0.13 & 0.56 & 0.49 & 0.12 & 0.55 & 0.64 & 0.14 \\ \bottomrule
\end{tabular}}
\end{table}
Table \ref{tab:res_util} confirms that the performance constraint, which mandates a maximum decrease in training task loss relative to the original model, also holds for the test loss across various datasets. Notably, our results reveal that incorporating the LLM-based rule refinement step leads to increased rule utilization while simultaneously reducing overall coverage in our experiments. This effect is particularly true for the MoRE approach with LR models. This strongly suggests that the knowledge alignment introduced by the LLM has a positive regularization effect. Specifically, it prunes rules that do not align with domain knowledge and increases the quality of the remaining rules.
The significant reduction in the number of rules due to rule pruning is shown in Table \ref{tab:res}. The performance comparison in the table demonstrates that MoRE-LLM outperforms white-box rule learning methods and is on par with non-interpretable tree ensemble methods and the MLP.
\begin{table}[]
\centering
\caption{Comparison of accuracy (acc.) and number of used rules between MoRE with and without LLM after three iterations and a selection of baselines. The methods are sorted by the complexity of interpreting the decision process.}
\label{tab:res}
\resizebox{\columnwidth}{!}{
\begin{tabular}{@{}cllllllll@{}}
\toprule
\multicolumn{1}{c}{\multirow{2}{*}{Method}} &
  \multicolumn{2}{c}{adult} &
  \multicolumn{2}{c}{g-credit} &
  \multicolumn{2}{c}{diabetes} &
  \multicolumn{1}{l}{\multirow{2}{*}{}} &
   \\ \cmidrule(lr){2-7} \cmidrule(l){9-9} 
\multicolumn{1}{c}{} &
  \multicolumn{1}{l}{acc.} &
  \multicolumn{1}{l}{\#Rules} &
  \multicolumn{1}{l}{acc.} &
  \multicolumn{1}{l}{\#Rules} &
  \multicolumn{1}{l}{acc.} &
  \multicolumn{1}{l}{\#Rules} &
  \multicolumn{1}{l}{} &
   \\ \cmidrule(r){1-8}\morecmidrules\cmidrule{1-2}\cmidrule(r){1-8}
\multicolumn{1}{c}{RIPPER} &
  \multicolumn{1}{l}{0.80} &
  \multicolumn{1}{l}{2} &
  \multicolumn{1}{l}{0.70} &
  \multicolumn{1}{l}{3} &
  \multicolumn{1}{l}{0.71} &
  \multicolumn{1}{l}{2} &
  \multicolumn{1}{c}{\multirow{2}{*}{Simple}} &
   \\ \cmidrule(r){1-7}
\multicolumn{1}{c}{CART} &
  \multicolumn{1}{l}{0.82} &
  \multicolumn{1}{l}{94} &
  \multicolumn{1}{l}{0.67} &
  \multicolumn{1}{l}{106} &
  \multicolumn{1}{l}{0.67} &
  \multicolumn{1}{l}{80} &
  \multicolumn{1}{c}{} &
   \\ \cmidrule(r){1-7}\morecmidrules\cmidrule{1-2}\cmidrule(r){1-7}
\multicolumn{1}{c}{MoRE (LR)} &
  \multicolumn{1}{l}{0.82} &
  \multicolumn{1}{l}{23} &
  \multicolumn{1}{l}{0.77} &
  \multicolumn{1}{l}{20} &
  \multicolumn{1}{l}{0.76} &
  \multicolumn{1}{l}{21} &
  \multicolumn{1}{c}{\multirow{4}{*}{\large$\downarrow$}} &
   \\ \cmidrule(r){1-7}
\multicolumn{1}{c}{MoRE-LLM (LR)} &
  \multicolumn{1}{l}{0.82} &
  \multicolumn{1}{l}{15} &
  \multicolumn{1}{l}{0.75} &
  \multicolumn{1}{l}{9} &
  \multicolumn{1}{l}{0.78} &
  \multicolumn{1}{l}{8} &
  \multicolumn{1}{l}{} &
   \\ \cmidrule(r){1-7}
\multicolumn{1}{c}{MoRE (MLP)} &
  \multicolumn{1}{l}{0.85} &
  \multicolumn{1}{l}{21} &
  \multicolumn{1}{l}{0.74} &
  \multicolumn{1}{l}{21} &
  \multicolumn{1}{l}{0.76} &
  \multicolumn{1}{l}{21} &
  \multicolumn{1}{l}{} &
   \\ \cmidrule(r){1-7}
\multicolumn{1}{c}{MoRE-LLM (MLP)} &
  \multicolumn{1}{l}{0.84} &
  \multicolumn{1}{l}{15} &
  \multicolumn{1}{l}{0.69} &
  \multicolumn{1}{l}{13} &
  \multicolumn{1}{l}{0.76} &
  \multicolumn{1}{l}{10} &
  \multicolumn{1}{l}{} &
   \\ \cmidrule(r){1-7}\morecmidrules\cmidrule{1-2}\cmidrule(r){1-7}
\multicolumn{1}{c}{RF} &
  \multicolumn{1}{l}{0.83} &
  \multicolumn{1}{l}{-} &
  \multicolumn{1}{l}{0.77} &
  \multicolumn{1}{l}{-} &
  \multicolumn{1}{l}{0.77} &
  \multicolumn{1}{l}{-} &
  \multicolumn{1}{c}{} &
   \\ \cmidrule(r){1-7}
\multicolumn{1}{c}{AdaBoost} &
  \multicolumn{1}{l}{0.82} &
  \multicolumn{1}{l}{-} &
  \multicolumn{1}{l}{0.72} &
  \multicolumn{1}{l}{-} &
  \multicolumn{1}{l}{0.75} &
  \multicolumn{1}{l}{-} &
  \multicolumn{1}{c}{} &
   \\ \cmidrule(r){1-7}
\multicolumn{1}{c}{GBDT} &
  \multicolumn{1}{l}{0.82} &
  \multicolumn{1}{l}{{-}} &
  \multicolumn{1}{l}{0.76} &
  \multicolumn{1}{l}{-} &
  \multicolumn{1}{l}{0.75} &
  \multicolumn{1}{l}{-} &
  \multicolumn{1}{c}{\multirow{4}{*}{Complex}} &
   \\ \cmidrule(r){1-7}
\multicolumn{1}{c}{MLP} &
  \multicolumn{1}{l}{0.85} &
  \multicolumn{1}{l}{-} &
  \multicolumn{1}{l}{0.75} &
  \multicolumn{1}{l}{-} &
  \multicolumn{1}{l}{0.77} &
  \multicolumn{1}{l}{{-}} &
  \multicolumn{1}{c}{} &
   \\ \bottomrule
\end{tabular}}
\end{table}
The results show that MoRE-LLM provides significantly simpler interpretations for parts of the input space, while matching the performance of non-interpretable approaches. Furthermore, considering the presented qualitative results, showing that the LLM makes reasonable adjustments to the rules to align them with domain knowledge, and the quantitative measurement of rule usage, it is clear that MoRE-LLM produces enhanced domain knowledge-aligned predictions.
\section{Conclusion}
We have introduced a framework to exploit the vast general knowledge inherent in LLMs to guide small task-specific grey-box models. We have shown that our MoRE-LLM approach can offer similar predictive performance as non-interpretable baselines and outperform interpretable white-box models while being better aligned with human domain knowledge and offering high fidelity rule-based explanations. As part of our method, we have demonstrated how LLMs can make valuable adaptations to logical rules and offer additional context to augment explanations. This insights might offer impulses for future research beyond this work.

\bibliographystyle{IEEEtran}
\bibliography{IEEEabrv,references}

\end{document}